\documentclass{article}

\usepackage[preprint]{neurips_2026}

\usepackage[utf8]{inputenc} % allow utf-8 input
\usepackage[T1]{fontenc}    % use 8-bit T1 fonts
\usepackage{hyperref}       % hyperlinks
\usepackage{url}            % simple URL typesetting
\usepackage{booktabs}       % professional-quality tables
\usepackage{amsfonts}       % blackboard math symbols
\usepackage{nicefrac}       % compact symbols for 1/2, etc.
\usepackage{microtype}      % microtypography
\usepackage{xcolor}         % colors
\usepackage{todonotes}
\usepackage{dsfont}
\usepackage{amsmath}
\usepackage{wrapfig}

\title{Mitigating Cognitive Bias in RLHF\\by Altering Rationality}
\author{
  Tiffany Horter \\
  University of Oxford\\
  \And 
  Andrew Markham \\
  University of Oxford\\
  \And
  Niki Trigoni \\
  University of Oxford\\
  \And
  Serena Booth \\
  Brown University \\
}

\begin{document}

\maketitle

%% The abstract must be limited to one paragraph.
\begin{abstract}
  How can we make models robust to even imperfect human feedback? In reinforcement learning from human feedback (RLHF), human preferences over model outputs are used to train a reward model that assigns scalar values to responses. Because these rewards are inferred from pairwise comparisons, this learning depends on an assumed relationship between latent reward differences and observed preferences, typically modeled using a Boltzmann formulation in which a rationality parameter $\beta$ informs how consistently preferences reflect reward differences. In practice, $\beta$ is typically treated as a fixed constant that reflects assumed uniform annotator reliability. However, human feedback is not this simplistic in practice: real human judgments are shaped by cognitive biases, leading to systematic deviations from reward-consistent behavior that arise contextually. To address this, we treat rationality as context- and annotation-dependent. We design an approach to dynamically adjust the rationality parameter $\beta$ during reward learning using an LLM-as-judge to assess the likely presence of cognitive biases. This approach effectively downweights comparisons that are likely to reflect biased or unreliable judgments. Empirically, we show that this approach learns a more rational downstream model, even when finetuning on datasets with strongly biased preferences. 
  \end{abstract}

\section{Introduction}
The underlying premise of reinforcement learning from human feedback, or RLHF, is that human annotators express their preferences correctly; this underpins this technique's widespread use in the alignment of large language models (LLMs). This premise is unfortunately flawed: human feedback is subject to the influence of cognitive biases~\cite{dalonzo2026helpful}.

A famous example of cognitive bias is the Linda problem \cite{TverskyAmos1983Evir}, a conjunction fallacy. People are told, ``Linda is 31 years old, single, outspoken, and very bright. She majored in philosophy. As a student, she was deeply concerned with issues of discrimination and social justice, and also participated in anti-nuclear demonstrations'' They must then assess whether A: ``Linda is a bank teller'' or B: ``Linda is a bank teller and is active in the feminist movement'' is more probable. Most people pick B despite the statistical impossibility of any subset of A being greater than the entire set of A. 

The Linda problem may seem trivial, but cognitive biases arise often in high-stakes domains such as medicine. For example, physicians are known to exhibit anchoring bias. This can lead a physician to potentially overweight an initial diagnosis (e.g., asthma) and discount subsequent evidence that may point to a more serious condition, like heart failure. These are not hypothetical concerns: empirical studies document systematic cognitive biases in clinical decision-making with real consequences for patient outcomes~\cite{ke2024medicalCogBias}. When cognitive biases are present in the judgments used as training signals, as in RLHF, these systematic deviations become embedded in the learned reward model, leading downstream models to reproduce or even amplify these biases in their outputs.

\begin{figure}[t]
    \centering
    \includegraphics[width=1.0\linewidth]{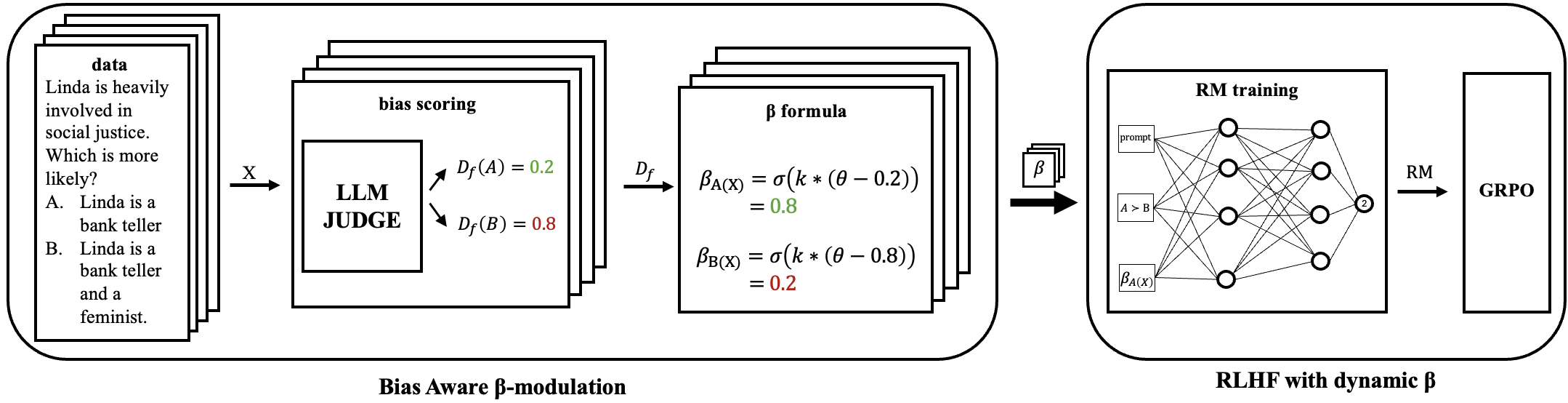}
    \caption{Overview of the intervention pipeline to reduce cognitive bias in the finetuned LLM. Humans provide preferences over responses to a prompt, like the ``Linda is a bank teller'' example. An LLM then judges whether each paired prompt and response is likely to be subject to a cognitive bias ($D_f$). From this measure, we compute a dynamic value of $\beta$. These prompts, responses, and $\beta$ values are used to learn a reward model and to finetune an LLM. Higher bias = lower value of $\beta$.} \vspace{-0.3cm}
    \label{fig:overview}
\end{figure}

To address the risk of cognitive biases in preference data, we intervene directly on the RLHF objective. In the standard formulation, human preferences are modeled using a Boltzmann-rational model with a rationality parameter $\beta$, which is typically fixed across annotators and across comparisons. This implicitly assumes uniform reliability in human judgments. We relax this assumption by treating $\beta$ as \emph{feedback-dependent}, as some responses are more likely to be unreliable than others. To do so, we estimate the likelihood that a given comparison is affected by cognitive bias, and we use this estimate to dynamically adjust $\beta$ during reward learning to downweight feedback that is likely to be biased; see overview Figure \ref{fig:overview}. This preserves informative signals from human preferences while reducing the influence of systematic cognitive biases, enabling the learned reward model and downstream policy to better reflect underlying preferences rather than observed, potentially biased judgments. 

We evaluate our approach on two datasets designed to elicit cognitive bias. Empirically, we find that dynamically adjusting $\beta$ by assessing response bias propensity yields the following:
\begin{enumerate}
\item \textbf{Reduction in bias propagation.} Relative to baseline LLMs without this intervention, our method produces models that are significantly less likely to prefer biased responses. In pairwise evaluations where one response exhibits a cognitive bias and the other does not, the finetuned model more frequently selects the unbiased response.
\item \textbf{Robustness to biased feedback.} Our approach mitigates a failure of RLHF under biased supervision: collapse toward biased preferences. By downweighting bias-prone preferences, the method remains effective even when a large fraction of training data reflects systematically biased annotations.
\item \textbf{Preservation of general performance.} Despite operating on heavily biased training data, the intervention does not degrade performance on unrelated tasks, indicating that the method selectively reduces bias without sacrificing overall model capability.
\end{enumerate}

\section{Background}
\subsection{Cognitive bias}
\label{sec:cog_bias}
RLHF has emerged as a prominent technique for aligning LLMs with human preferences. This is often framed as progress toward ``value alignment,'' the challenge of ensuring that learned objectives faithfully reflect human intent \cite{russellHumanCompatibleAI2019}. However, RLHF adopts the premise that expressed human preferences reflect human intentions. But, humans are fallible: we regularly make mistakes and are subject to cognitive biases, so our expressed preferences may not capture our intentions. Standard RLHF pipelines do not differentiate between more or less reliable forms or instances of feedback.

Similar to the problem of reward misspecification in reinforcement learning, where even well-intentioned designs can yield unintended behavior, RLHF can inherit systematic irrationality from human judgment. One contributing factor, consistent with prior work on reward design~\cite{boothPerilsTrialanderrorReward2023a}, is that human annotators rely on simplified, often myopic reasoning when expressing preferences. Annotators may neglect longer-term consequences or cumulative effects and instead rely on heuristics. For example, prospect theory shows that humans are more sensitive to relative losses than to gains of the same magnitude, leading to asymmetric evaluations of outcomes~\cite{kahneman2013prospect}, which may manifest in preferences. Likewise, temporal preferences are often inconsistent with exponential discounting; people exhibit hyperbolic discounting and disproportionately favor immediate rewards over delayed ones \cite{moore2025and}. Additional biases such as framing effects, anchoring, and scope insensitivity can further shape expressed preferences in ways that are not aligned with stable or reflective intentions~\cite{dalonzo2026helpful,hatgis2025influencing}.

% Some of these forms of irrationality in annotator preferences may be unpredictable, but some of them can be anticipated and compensated for. 
For humans, cognitive biases serve a purpose: they enable rapid decision-making through heuristics, trading off accuracy for speed. There is an established literature that shows humans are more likely to experience cognitive biases under certain conditions and therefore act in a less rational way \cite{macmillan-scottIrrationalityAIState2023, kahnemanThinkingFastSlow2013}, e.g., under time pressure. Because LLMs are trained with human preferences, LLMs tend to adopt our cognitive biases during pretraining \cite{malbergComprehensiveEvaluationCognitive2024} or in finetuning whether through instruction tuning or RLHF ~\cite{cheungLargeLanguageModels2025, itzhakInstructedBiasInstructiontuned2024}. Unlike humans, however, LLMs are not subject to the same time and resource constraints that encourage the use of such heuristics. This creates an opportunity: rather than inheriting human biases as an unavoidable feature of decision-making or RLHF, we can design learning procedures that identify and mitigate them, enabling LLMs to better approximate deliberative reasoning. 
%Standard LLM architectures apply a fixed amount of computation to each query, reducing the benefits of biased, fast reasoning while preserving the risks of bias if learned from biased data.

In decision-aid settings, there is a fundamental tension between matching human-provided feedback and avoiding the systematic biases that such feedback may contain. In this work, we assume that people hold true preferences that may be obscured in their expressed preferences when cognitive bias is likely present \cite{hosking2024humanfeedbackgoldstandard}. We seek to learn the underlying preferences that a person would express were it not for the confounding factors of the bias. Others have shown that under some circumstances, people prefer machines not act in accordance with the literal preference given but rather with the intent behind the wording~\cite{horterBoothShouldRobotsComply2026}. In a similar vein, we assess the level of rationality a human annotator is likely to be experiencing when providing preferences over prompts and responses, and scale weight of this preference into the reward model approximation accordingly. This can in part address two known open problems with RLHF~\cite{casperOpenProblemsFundamental2023}, namely (A) ``Humans make simple mistakes due to limited time, attention, or care'' and (B) ``Humans can be misled, so their evaluations can be gamed.'' 

\subsection{Boltzmann Rationality}
The premise of this work (and of RLHF more broadly) is that people have a latent reward function $r^*$ that induces a distribution over their preferences, or that this is a reasonable modeling assumption. The objective of RLHF is to approximate this reward function from expressed preferences. The standard model of human decision-making uses a Boltzmann-rational assumption~\cite{jeonRewardrationalImplicitChoice2020a}. Under this model, humans are more likely to select higher-reward choices, with the strength of this tendency increasing as reward differences grow. In this model, there is a parameter $\beta$ that is known as the rationality parameter, and it controls how consistently higher-reward responses are selected.  

Although our proposed approach of intervening on the rationality parameter could apply more broadly to settings that learn from expressed human choices (e.g., learning from demonstrations), we focus on the RLHF setting, where a reward function is inferred from pairwise preferences over model outputs. In particular, we condition on a candidate reward function r and model the likelihood of observed comparisons between outputs. Given a comparison $\sigma_1 \succ \sigma_2$, this yields the standard logistic form:
\[
\mathds{P}(\sigma_1 \succ \sigma_2 \mid r)
= \texttt{logistic}\left(\beta \left(r(\sigma_1) - r(\sigma_2)\right)\right)
\]
In standard approaches of approximating a reward function, whether from preferences or other modes of human feedback, $\beta$ is treated as a fixed parameter capturing the overall noisiness of human feedback. This interpretation assumes that deviations from reward-consistent behavior arise from uniform stochastic noise, rather than systemic bias of some form.
%In this formulation, the log-odds of preferring $\sigma_1$ over $\sigma_2$ are linear in the reward difference, with slope $\beta$.

We instead treat the mapping from latent preferences to expressed preferences as context-dependent. Rather than assuming a fixed $\beta$, we allow $\beta$ to vary across instances based on the expected reliability of the feedback. Intuitively, $\beta$ should be lower in settings where cognitive biases are likely to distort judgment, and higher when responses are more likely to align with latent preferences. This can be viewed as reweighting observations by their expected fidelity, with the goal of recovering a reward model that better approximates the true $r^*$, rather than fitting the reward model to artifacts of bias.

\subsection{Prior Work on Rationality Modeling}

Accurately modeling the human’s true decision process, even under cognitive bias, is critical for effective reward learning \cite{hong2023sensitivityrewardinferencemisspecified, knoxModelsHumanPreference2023,chan2021humanirrationalitybadgood}. In fact, explicitly modeling structured human irrationality can improve reward inference beyond assuming perfectly rational behavior, even with an oracle \cite{chan2021humanirrationalitybadgood}.

Prior methods have explored how rationality varies with feedback modality (e.g., comparisons, demonstrations, corrections) \cite{ghosalEffectModelingHuman2022}, as well as with annotator expertise \cite{daniels-kochExpertiseProblemLearning2022}. Other work addresses heterogeneity in annotator rationality by computationally varying the $\beta$ parameter \cite{yamagataRelativelyRational2024, barnettActiveRewardLearning2023}, though these approaches differ from our approach as they do not model bias as context-dependent or as a property of the response. Some approaches instead fit a global $\beta$ to account for the level of systematic bias in human responses across feedback types~\cite{ghosalEffectModelingHuman2022}, whereas we focus on the information held in the response rather than its format. Others attempt to learn the algorithms people use to make decisions directly, often finding that incorporating known heuristic biases improves learning performance \cite{shahFeasibilityLearningRather2019}.

More recent work adjusts $\beta$ based on estimated difficulty of the annotation setting, such as interaction signals (e.g., clicks, time spent) or model-predicted difficulty \cite{singhalScalableOversightAccounting}. However, these approaches treat rationality as a function of the overall scenario rather than the specific feedback provided over responses, and do not evaluate whether such adjustments mitigate cognitive bias, focusing instead on upstream reward model behavior (e.g., sensitivity to factuality versus length). The issue with accounting for only the scenario rather than the response is that it removes or downweights \emph{all} responses to biased scenarios, and ignores responses that avoid the pitfall of biases. We suspected that conditioning bias on the scenario alone would lead to suboptimal performance; see Appendix \ref{app:beta-formula-ablation-scenario}.

% The Boltzmann-rational model has many known issues~\cite{lindner2022humansboltzmanndistributionschallenges}. We address we some of these by exposing LLM finetuning to data that induces cognitive biases and mitigating this bias by treating it as context-dependent.

\section{Method}
\label{sec:method}
We build on the RLHF formulation introduced above, where the rationality parameter $\beta$ governs how strongly reward differences influence observed preferences. Rather than treating $\beta$ as fixed, we model this parameter as instance-dependent, reflecting the reliability of each piece of feedback.

Given an observed comparison, we estimate whether the expressed preference is likely to be influenced by cognitive bias. To do so, we use an out of the box LLM-based evaluator that takes as input the prompt, candidate responses, and annotation, and predicts whether the observed preference reflects a biased judgment.

We then adjust $\beta$ accordingly. When feedback is likely to be affected by bias, we reduce $\beta$ to downweight its influence; when feedback is likely to reflect the annotator’s latent preference, we increase $\beta$ to place greater weight on the observation. In this way, $\beta$ acts as a context-dependent measure of how faithfully expressed preferences reflect latent preferences.

% This differs from prior work that varies $\beta$ based on annotator identity or global properties of the task. Instead, our approach conditions directly on the observed feedback and its susceptibility to bias.
% Given these instance-dependent $\beta$ values, we train a reward model using the modified likelihood and subsequently use this reward model to finetune the policy via GRPO.

This proposed dynamic, instance-dependent rationality parameter is defined as
\begin{equation}
\beta_{new} = \texttt{logistic}\!\left(k \cdot (\theta - D_f)\right),
    \label{eqn:b_new_main}
\end{equation}
where $D_f \in [0,1]$ denotes the estimated probability that the observed feedback is influenced by cognitive bias, $\theta$ is a threshold parameter, and $k$ controls the steepness of the transition. This parameterization maps bias likelihoods to a continuous measure of feedback reliability. When $D_f$ is low (i.e., the feedback is unlikely to be biased), $\beta_{new}$ approaches 1, placing greater weight on the observed preference. As $D_f$ increases, $\beta_{new}$ decreases smoothly toward 0, reducing the influence of potentially biased feedback. The use of a logistic transform ensures a bounded mapping, while allowing for a tunable transition around the threshold $\theta$. This corresponds to human reasoning: we don't update as strongly on information we deem to be likely to not be trustworthy or to be biased. 

The hyperparameters $k$ and $\theta$ control the sensitivity of this mapping. The threshold $\theta$ determines the bias likelihood at which feedback begins to be substantially downweighted, while $k$ governs how sharply this transition occurs. In practice, we select these parameters via a 2-D traversal of the validation set, selecting values that maximize the model’s ability to distinguish between ground-truth-consistent and cognitively biased preferences. These hyperparameters may shift across datasets or annotator populations and reflect how susceptible to bias the annotator population is likely to be.

Importantly, $D_f$ is defined with respect to the \emph{human's chosen response} rather than the scenario as a whole. This distinction allows the model to differentiate between contexts that are conducive to bias where annotators fall into those biases and instances in which annotators nevertheless provide reliable feedback. When a scenario is bias-inducing but the annotator prefers an unbiased response, the corresponding $D_f$ remains low, and the feedback is assigned a higher $\beta_{new}$. Such a scenario might occur in someone with statistical training who encounters the Linda example, who would be primed to recognize that having a larger subset than the original set is impossible. In the reverse case, we have situations like the patient with heart failure who presents to the hospital with chest pain. In this example, the context is not primed to trigger cognitive bias, but the doctor's response only mentions having seen asthma on the chart, indicating a likely anchoring bias. Considering the bias of the chosen response avoids discarding informative examples from difficult or ambiguous settings, and instead treats such cases as high-value signals for learning under challenging conditions.

% Overall, this formulation treats $\beta$ as a continuous, instance-level measure of annotation reliability, enabling the model to downweight systematically biased feedback while preserving informative signals from complex or bias-prone contexts.

\subsection{Data}
\label{data-explanation}
To validate this intervention, we used two datasets designed to induce biased responses. The first is the BRU Dataset \cite{zhong2025balancing} which was developed with a psychologist and a medical data expert to ensure validity. This small dataset consists of 205 multiple choice questions with a ground truth answer and covers 8 cognitive biases (Anchoring Bias, Base Rate Fallacy, Conjunction Fallacy, Gambler’s Fallacy, Insensitivity to Sample Size, Overconfidence Bias, Regression Fallacy, Sunk Cost Fallacy). The second dataset is the ``Comprehensive Evaluation of Cognitive Biases in LLMs: Dataset'' (CogBias dataset~\cite{malberg-etal-2025-comprehensive}) that includes 30,000 choice selection test cases and 30 biases~\cite{malberg-etal-2025-comprehensive}. Stereotyping bias examples were excluded as they reflect social bias rather than a cognitive bias. This dataset was designed only for testing models for bias; we received special permission from the authors to use the finetune with this dataset to proactively reduce replicated cognitive bias in models. 

Across both datasets, to be consistent with RLHF framing, we convert each example from the initial multiple choice options into a binary preference pair consisting of a ground-truth response and a bias-consistent response. The former aligns with the unbiased or intended evaluation of the task; this is the preference a human might arrive at if they employed deliberative reasoning instead of heuristic-driven reasoning~\cite{guthrie2007blinking,kahnemanThinkingFastSlow2013}. The latter reflects the direction of error induced by a cognitive bias: for example, in the Linda problem, selecting an example that exhibits the conjunction fallacy (that it is more likely she is a bank teller and active in the feminist movement). Since neither dataset contained train/val/test splits, we split the data into train, val and test (shared in code).

In the BRU dataset, the unbiased choice ground-truth labels are provided. We construct bias-consistent responses by prompting a language model to select the response most aligned with a specified bias type. In the CogBias dataset, we use the provided bias metric to select the least biased option as the ground-truth-consistent response and the most biased option as the bias-consistent response.

To evaluate robustness to biased annotations, we simulate systematic bias in human feedback by constructing a dataset that consists of preference pairs where the biased response is more often (but not universally) preferred over the ground truth answer. Specifically, we construct this dataset by marking the bias-consistent response as preferred over the ground-truth-consistent response a tunable fraction $(> 0.5)$ of the time. This models a common failure mode in RLHF, where annotators systematically favor more salient or cognitively appealing options rather than the correct ones.

By default, we use a 3:1 ratio of bias-consistent to ground-truth-consistent labels in our experimentation, reflecting settings in which biased responses are more likely but not universal. We additionally vary this ratio from 1:1 to 5:1 in Section~\ref{pair_corruption} to study sensitivity to the level of bias.

\subsection{Bias detector}
The first step of our method is to estimate, for each training example, the likelihood that the observed feedback is influenced by cognitive bias. This estimate, denoted $D_f$, is used to modulate the contribution of each example during reward learning.

This approach uses an LLM-based evaluator to estimate the likelihood that a given preference comparison reflects cognitive bias. Here, we frame bias detection as a local judgment over a single comparison reflecting its contextual basis rather than a consistent aggregation across many examples. As we show empirically, the method remains effective even when the bias detector is imperfect, indicating that even coarse estimates of bias likelihood are sufficient to guide the intervention. 

To demonstrate that our approach is compatible with different bias detection strategies, we instantiate $D_f$ using two LLM-based evaluators. For the BRU dataset, we use ChatGPT 5.2 Thinking~\cite{openai2025gpt52systemcard} to score the likelihood that a given preference reflects the influence of cognitive bias. For the CogBias dataset, since the dataset is much larger, we used a local model as the evaluator, Mistral-7B-Instruct \cite{jiang2023mistral7b}. We construct $D_f$ using a pairwise LLM-as-judge procedure. Given two candidate responses, the model is prompted to identify which response is more biased. To mitigate positional bias, we evaluate each pair in both orderings and aggregate the results. We then compute a probability by applying a softmax over the final-token logits of the two choices.

Because bias detection performance varies across bias types (i.e., a model may be able to reliably detect anchoring bias but not conjunction fallacies), we apply a lightweight calibration step. Specifically, we estimate a per-bias-type transformation using a minimal annotated held-out validation set (20 examples per bias) to correct systematic directionality errors via sign inversion. Importantly, the bias detector is not trained with ground-truth correctness labels. This design reflects a realistic setting in which large-scale bias annotations are unavailable, while still enabling effective estimation of $D_f$.

These two instantiations provide evidence that the approach is not tightly coupled to a particular bias detection mechanism. Despite differences in model scale and prompting strategy, both yield effective estimates of $D_f$, suggesting the method is robust to variation in how likelihood is operationalized.

\subsection{Reward Model \& GRPO Finetuning}
\label{reward_model}
Given the training set consists of a mix of groundtruth-consistent and bias-consistent preference pairs that simulate cognitively-biased annotation, we train a reward model with a loss function that treats each pair's weight as conditional on how biased the chosen response appears to be. Concretely, we replace the global Bradley–Terry rationality constant $\beta$ with a per-datapoint weight $\beta_i$ that decays smoothly when the chosen response's bias score $D_f$ is high. 

The reward model is initialized from Mistral-7B-v0.1 \cite{jiang2023mistral7b}. The $\theta$ and $k$ parameters are selected via a two dimensional sweep to maximize accuracy over the validation set. The final $k, \theta$ values per dataset, as well as reward model training hyperparameters are in Appendix \ref{app:cogbias-training-hyperparams} and \ref{app:bru-training-hyperparameters}.      
The models were then finetuned with Group Relative Policy Optimization (GRPO) \cite{shao2024deepseekmathpushinglimitsmathematical}, where rewards were determined by a frozen reward model. All hyperparameters were held identical across runs for the baselines and debiased models. Our standard baseline for the finetuning was to use $\beta=1.0$ \cite{barnettActiveRewardLearning2023,ghosalEffectModelingHuman2022}. This simulates annotators being equally rational in all scenarios. We also tested a variety of fixed $\beta$ values (0.1, 0.5, 0.9) and a random $\beta$  to confirm that there was not a hidden fixed $\beta$  value that was best or that per-example modulation was the key to the improvement mechanism.

Testing was conducted on a held out test split from each dataset. We generated $n$ samples per prompt in the test set to reduce dependence on a random draw. These answers were then compared to the ground truth answers and those that matched the ground truth were marked as correct. Our main metric to assess the performance improvements of dynamic $\beta$ is the accuracy (the direct ground-truth rate) since it is RM independent. A significance level of $\alpha = 0.05$ was used globally. 

\section{Evaluation on the BRU \& CogBias Datasets}
We evaluate the performance of this method on two datasets. One is a small scale study on the BRU dataset (205 questions total, 21 in the test split) to validate the promise of this method. The other is the CogBias dataset (29,000 test cases). Across these experiments, we found support for the hypothesis that dynamic $\beta$ selection can reduce an LLM's replication of cognitive biases.

\subsection{Reduction in bias propagation \& preservation of general performance}
\begin{figure}[t]
    \centering
    \includegraphics[width=0.8\linewidth]{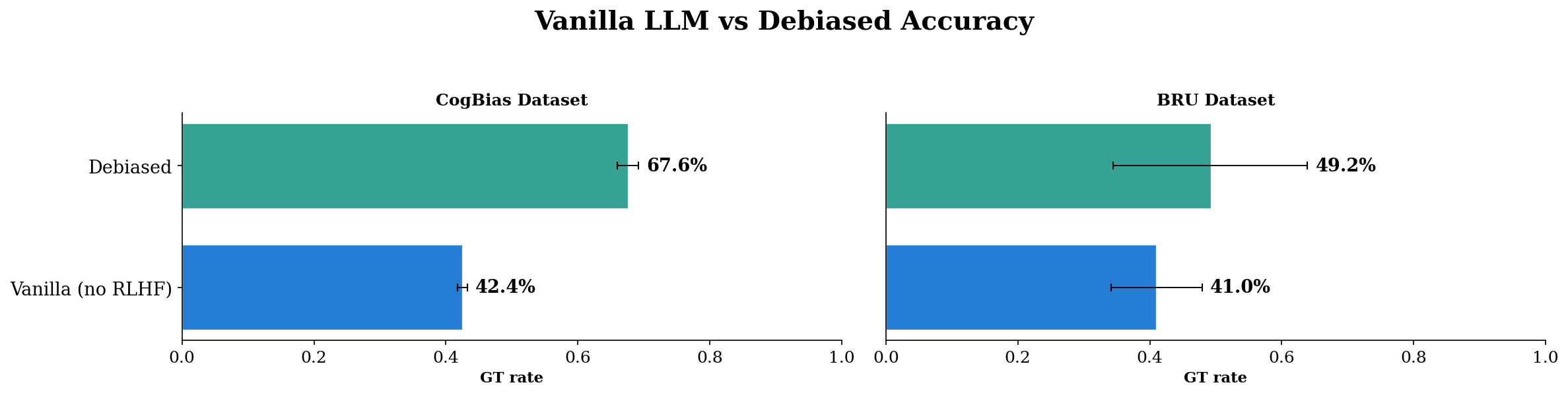}
    \caption{For CogBias (left), the debiased model's accuracy of $67.6\%$ correctly choosing ground truth is a significant improvement over the vanilla un-finetuned LLM with ground truth rate $42.4\%$. } \vspace{-0.3cm}
    \label{fig:compare-vanilla}
\end{figure}

To investigate whether the intervention could reduce bias propagation, we compared the performance of the debiased model with the base model. Despite exposure to a substantial number of biased samples, the debiased model still outperforms the base model which was never exposed to the biased data on the both the test set of the CogBias and BRU dataset. On the BRU dataset, the debiased model improves by $8.3\%$ over the base vanilla (untrained) LLM despite having been finetuned on $3:1$ cognitively biased data, though it doesn't reach $\alpha < 0.05$ with $n=21$ prompts. The CogBias results show significant improvement of $25.2\%$ ($95\%$ CI $[+23.3, +26.9]$, $d = 0.51$). This lends support to our claim of a reduction in bias propagation through this dynamic $\beta$ intervention. See figure~\ref{fig:compare-vanilla}. 

Further, this increase in performance on cognitive bias tasks does not come at the cost of general performance. To test this, we compared the vanilla LLM and the debiased model (trained with the CogBias dataset and the dynamic $\beta$ intervention) on the TruthfulQA dataset~\cite{lin2022truthfulqameasuringmodelsmimic}, and found it did not significantly degrade performance (non-significant change of $-0.5\%$ with $95\%$ CI $[-0.011, 0.000]$).

\subsection{Robustness to biased feedback}
\begin{figure}[b]
    \centering
    \includegraphics[width=\linewidth]{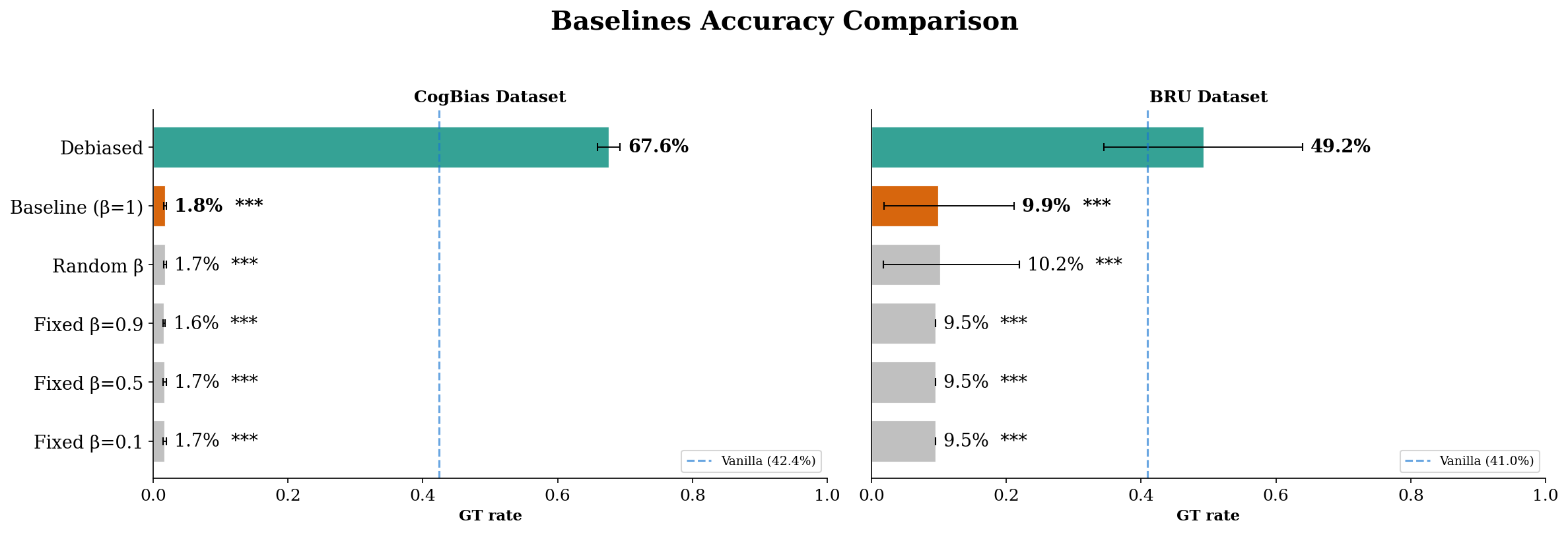}
    \caption{Comparison of CogBias (left) and BRU (right) datasets accuracy compared across different baselines ($\beta \in \{1, 0.9, 0.5, \texttt{random}\}$) and our dynamic $\beta$  method (Debiased, top row). Our method shows a strong improvement in accuracy compared to the baselines. With either fixed or random choices of $\beta$, these models were highly susceptible to cognitive biases.}
    \label{fig:baselines-accuracy}
\end{figure}
The performance on both datasets shows that adding a model of cognitive bias to the reward model tuning makes it robust to even high levels of human error due to cognitive bias; see Figure \ref{fig:baselines-accuracy}. When finetuned on a dataset where each question had three biased and one unbiased responses to model the distribution of cognitive bias, we found that our debiased method (67.6$\%$ accurate for CogBias, 49.2$\%$ for BRU) far outperformed the baselines (Cohen's d >= 1.48 for CogBias) that trained on the same data but with fixed values for $\beta \in \{0.1, 0.5, 0.9, 1.0\}$ or where $\beta$ was assigned a random value for each question. All baselines produce GT rates roughly indistinguishable from the collapsed baseline $(\approx9-10\%$ for BRU dataset, $\approx1.7\%$ for CogBias dataset). The baselines' degradation in performance is expected, as each has been exposed to imperfect data and has learned to predict that imperfect data. Only our dynamic $\beta$ improves robustness with biased human data, with bootstrap $p<0.01$ for all baselines in both datasets (for BRU fixed-$\beta$: one-sided). This is promising, as the system learned to avoid biases without structured heuristics or privileged information about the bias. 
    
\subsection{Robustness to different levels of biased feedback}
\label{pair_corruption}
Our method is robust to different levels of bias in the dataset, as shown in Table \ref{tab:pair-ratio}. To study this, we finetuned with the baseline $\beta =1.0 $ and dynamic $\beta$ on training data with different ratios of preferring biased or unbiased responses. We trained both on fixed ratios (1:1, 3:1, 5:1) and on one dataset where the number of biased pairs was randomly selected for each question. In all cases, despite finetuning on preference pairs that did not match people's true underlying preferences, the debiased model still has higher accuracy in selecting more rational responses than the un-finetuned base model.

\begin{table}[t]
  \centering                                           
  \caption{CogBias: Comparing different ratios of ground-truth preferences to cognitively biased preferences by ground-truth answer rate, effect size, and reward-based head-to-head win rate. All values use $n{=}10$ samples per prompt over 2,900 test prompts. Win rate is reward-based: per-prompt mean debiased-RM reward, debiased vs. baseline. Intervention shows strong improvement across different levels of cognitive bias. BL GT: baseline ground truth. DB GT: debiased ground truth.}
  \label{tab:pair-ratio}                                       
  \begin{tabular}{rrrrrrr}
    \toprule                                            
    Ratio & BL GT\% & DB GT\% & GT diff & Cohen's $d$ & Win rate (rwd) & $p$-val \\
    \midrule                                         
     1:1                              & 47.9\% & 82.7\% & $+34.8\%$ & 0.81 & 90.1\% & $<\!0.01$ \\            
    \textbf{1:3}                     &  1.8\% & 67.6\% & $+65.8\%$ & 1.48 & 71.8\% & $<\!0.01$ \\ 
    1:5                              &  1.8\% & 53.5\% & $+51.7\%$ & 1.08 & 61.9\% & $<\!0.01$ \\   
    random:3                         & 49.3\% & 83.7\% & $+34.4\%$ & 0.64 & 66.0\% & $<\!0.01$ \\    
  \bottomrule     
  \end{tabular} \vspace{-0.4cm}          
\end{table}

\subsection{Robustness to model architectures}
We also tested to assess whether this intervention applied across model architectures. We tested with Mistral-7b-0.1~\cite{jiang2023mistral7b} and Qwen3-8B-Base~\cite{qwen3technicalreport}. Due to compute limitations, Qwen3-8B-Base was quantized to use a 4-bit policy quantization for GRPO; hyperparameters in Appendix \ref{app:robustness-hyperparam}. Across both models, the debiased method showed a statistically significant improvement (Cohen's $d$, Table \ref{tab:robustness}). 

\begin{table}[b]    %\vspace{-0.5cm}  
  \centering                                                                                       \small   
  \caption{Cross-architecture robustness on CogBias dataset. Each row is a base model finetuned with the beta modulation pipeline; OTB is the un-tuned base. All $n = 2900$ prompts $\times$ 10 samples. CIs from 10,000-iteration paired bootstrap; $p$-values from one-sided Wilcoxon signed-rank test ($H_1\colon \Delta > 0$); Cohen's $d$ on per-prompt paired differences.}              
  \label{tab:robustness}
  \begin{tabular}{lcccccc}                                                                        
    \toprule        
    Base model & OTB GT & Debiased GT & $\Delta$ (pp) & 95\% CI & Wilcoxon $p$ & Cohen's $d$ \\
    \midrule    
    \textbf{Mistral-7B-v0.1} & 42.4\% & 67.6\% & $+25.2$ & $[+23.3,\, +26.9]$ & $<\!0.001$ & $+0.509$ \\
    % OLMo-3-1025-7B           & 46.2\% & 50.4\% & $+4.2$  & $[+2.4,\, +6.0]$   & $<\!0.001$   & $+0.085$ \\ 
    Qwen3-8B-Base            & 45.8\% & 60.0\% & $+14.21$  & $[+12.5,\, +15.9]$   & $<\!0.001$  & $+0.31$ \\  
  \bottomrule                                                                                     
  \end{tabular}   \vspace{-0.8cm}                                                                         
  \end{table}  

\subsection{Robustness to noisy judging}
To confirm the bias detection worked on the BRU dataset, we tested the pairwise accuracy $(D_f(cb) > D_f(gt))$ of the ChatGPT 5.2-Thinking model; this was $98.0\%$ accurate. All 8 types of bias in this dataset scored above $93\%$ accuracy in detection and therefore did not require calibration.

On the CogBias dataset, at training time, the per-pair $\beta$ values were able to cleanly separate ground truth pairs from cognitive biased ones. On the held-out validation set, the calibrated out-of-the-box Mistral-7b-0.1-Instruction-Tuned \cite{jiang2023mistral7b} judge achieves $82.2\%$ pairwise accuracy (Cohen's $d = 0.696$) across bias types, with $24/29$ types significantly above chance (Wilcoxon $p < 0.001$).

Given the intervention relies on bias scores from a potentially noisy judge, it is important that the method can tolerate noise when the judge is wrong. To probe this, we perturbed the bias scores by adding zero-mean Gaussian noise in logit space to bias scores and recalculated $\beta$ on the $n=2900$ CogBias validation pairs with the $3:1$ ratio of $gt$ to $cb$. When the judge is correct $(D_f(cb) > D_f(gt))$, $\beta_{cb} < \beta_{gt}$ correctly downweights the $cb$-chosen pairs; when the judge is incorrect, it flips so $\beta_{cb} > \beta_{gt}$ incorrectly downweights the ground-truth-chosen pairs. We summarize this trade-off as the benefit/damage ratio: avg($\beta_{gt} - \beta_{cb}$) on judge-correct pairs divided by $avg(\beta_{cb} - \beta_{gt})$ on judge-flipped pairs. A ratio greater than 1.0 indicates that the $\beta$-modulation is helping more than it is harming. Across the judge accuracy (induced by the noise addition) from $83\%$ (no noise) down to $57\%$ (near chance), the benefit/damage ratio stays $> 1$ $(1.84 \rightarrow 1.12)$. The 1.0 failure threshold is not crossed; we conclude the intervention is therefore relatively insensitive to adjudication noise.

\subsection{Transferring learned knowledge of cognitive biases to other datasets}
Does the model learn a transferrable representation of the cognitive biases? We tested the debiased model trained on the BRU dataset on the subset of test prompts from the CogBias dataset that shared a common cognitive bias: anchoring bias (100 samples, randomly assigned equal numbers of A/B labels). We find that the BRU-trained debiased model is successfully able to transfer that knowledge of anchoring bias onto the new dataset, having a $44.3\%$ GT accuracy improving over the vanilla LLM's $37.7\%$ GT accuracy (p=0.001, paired Wilcoxon). This suggests that this method has promise for being used more broadly on datasets drawn from a different distribution. Given the small scale of the BRU dataset, it is surprising and encouraging that it was able to transfer structural knowledge. 

\section{Limitations}
\begin{wrapfigure}{r}{0.5\textwidth}
    \begin{center}
        \includegraphics[width=1.0\linewidth]{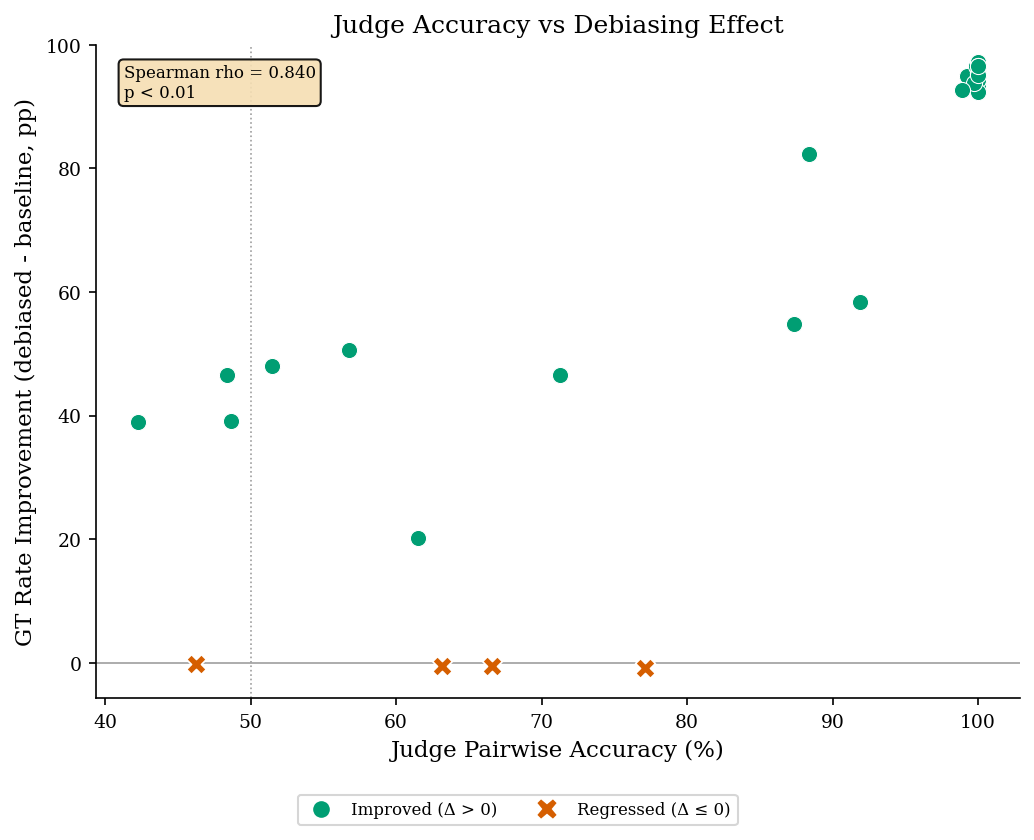}
    \end{center}
    \caption{Does more accurate judging (i.e., $D_f$) result in more effective debiasing? Judge accuracy compared to ground truth rate improvement (debiased - baseline). All low-performing ($\sim 0$ improvement) have judge accuracy under $80\%$. } \vspace{-0.5cm}
    \label{fig:biascatterdownstream}
\end{wrapfigure}

One limitation of this method is that the power of this transformation depends in part on the LLM-as-judge's ability to detect a given type of bias. In these experiments, we observed that the LLM was able to correctly assess some types of bias, while not performing well on others. We therefore probed whether model performance after finetuning could be attributed to how accurate the LLM judge was; we break this analysis down by bias types. For the CogBias dataset, all 21 bias types with a judge accuracy greater than $85\%$ improve by $>80$pp while the 4 with judge accuracy less than $55\%$ improved by $\leq 1$ pp; even with these effectively random judges, the performance did not degrade further. As shown in Fig.~\ref{fig:biascatterdownstream}, the Spearman $\rho$ of $0.84$ with $p < 0.01$ confirms the positive relationship between the judge accuracy and the debiasing effect, showing that a limiting factor on this method's success is whether the LLM-as-judge can recognize the bias when it occurs. In general, though, we expect that even if the judge were not a strong supervisor, it should show promising results, given that work has shown finetuning a strong model on labels from a weak model leads to improved generalized performance over the weak model~\cite{burns2023weaktostronggeneralizationelicitingstrong}. In addition, we expect that the task of determining if bias is present is part of the training corpus and therefore recoverable; whereas avoiding such a bias is not, as in \ref{sec:cog_bias}. There is promise in this method; even in the presence of an ineffective judge, our experiments did not yield performance regressions.
%it is easier for humans to detect bias than it is for them to correct for it; this is likely true of LLMs, which are trained to a large extent in mimicry. 

\subsection{Ethics}
There is, of course, always a moral dilemma when we work on the question of how can we reinterpret humans' expressed preferences: what does it mean to say that we are able to recover the true or accurate preference when that preference is not the one we express or choose? We run the risk of paternalism when we allow an LLM to determine truth despite explicit feedback. However, there is a key distinction here, as we are only applying this reinterpretation in situations where long-running psychological experiments have shown that humans make less rational choices. The biases and heuristics that may be helpful for humans to short circuit computations should not be unthinkingly passed on to an LLM that people will then use and be themselves be impacted by cognitive bias (i.e.,~confirmation bias) to accept its prescriptions and responses with little questioning. 

\section{Conclusion}
In this work, we revisit an old assumption in preference-based reward learning: that deviations from reward-consistent behavior can be captured by a fixed, global rationality parameter. We argue that this view is inadequate for real human feedback, where systematic biases and contextual effects shape expressed preferences in structured ways. By treating $\beta$ as annotation-dependent and using an LLM-based judge to estimate the likelihood of bias, our approach reframes noise as coming from a signal that can be in part identified and accounted for during training. 

Empirically, we show that this intervention improves downstream reward learning even when bias detection is imperfect, suggesting that robustness to human inconsistency does not require perfect models of cognition. Our results contribute to a growing direction: improving alignment may depend less on idealized feedback and more on modeling the distortions in feedback~\cite{chan2021humanirrationalitybadgood,ethayarajh2024kto,knox2024learning}.

More broadly, this work argues for a richer view of RLHF as a measurement process. Rather than collapsing human judgments into a single scalar signal under strong assumptions of rationality, future work should treat preference data as context-dependent, heterogeneous, and systematically biased. Developing methods that can detect, model, and adapt to these properties is critical for effective alignment, enabling reward models to leverage imperfect feedback without collapsing to its biases.

%% Papers to be submitted to NeurIPS 2026 must be prepared according to the
%% instructions presented here. Papers may only be up to {\bf nine} pages long, including figures. \textbf{Papers that exceed the page limit will not be reviewed (or in %% any other way considered) for presentation at the conference.}
%% Additional pages \emph{containing acknowledgments, references, checklist, and optional technical appendices} do not count as content pages. 

\bibliographystyle{plainnat}
\bibliography{AlterRationality.bib}

@misc{barnettActiveRewardLearning2023,
  title = {Active Reward Learning from Multiple Teachers},
  author = {Barnett, Peter and Freedman, Rachel and Svegliato, Justin and Russell, Stuart},
  year = 2023,
  eprint = {2303.00894},
  primaryclass = {cs.LG},
  archiveprefix = {arXiv}
}

@inproceedings{yamagataRelativelyRational2024,
title = "Relatively Rational: Learning Utilities and Rationalities Jointly from Pairwise Preferences",
author = "Taku Yamagata and Tobias Oberkofler and Timo Kaufmann and Viktor Bengs and Eyke H{\"u}llermeier and Raul Santos-Rodriguez",
year = "2024",
month = aug,
day = "23",
language = "English",
note = "ICML 2024 Workshop on Models of Human Feedback for AI Alignment ; Conference date: 23-08-2024 Through 23-08-2024",
}

@article{casperOpenProblemsFundamental2023,
  title = {Open Problems and Fundamental Limitations of Reinforcement Learning from Human Feedback},
  author = {Casper, Stephen and Davies, Xander and Shi, Claudia and Gilbert, Thomas Krendl and Scheurer, J{\'e}r{\'e}my and Rando, Javier and Freedman, Rachel and Korbak, Tomasz and Lindner, David and Freire, Pedro J and Wang, Tony and Marks, Samuel and S{\'e}gerie, Charbel-Rapha{\"e}l and Carroll, Micah and Peng, Andi and Christoffersen, Phillip J. K. and Damani, Mehul and Slocum, Stewart and Anwar, Usman and Siththaranjan, Anand and Nadeau, Max and Michaud, Eric J. and Pfau, Jacob and Krasheninnikov, Dmitrii and Chen, Xin and {di Langosco}, Lauro Langosco and Hase, Peter and Biyik, Erdem and Dragan, Anca D. and Krueger, David and Sadigh, Dorsa and {Hadfield-Menell}, Dylan},
  year = 2023,
  journal = {ArXiv},
  volume = {abs/2307.15217}
}

@article{cheungLargeLanguageModels2025,
  title = {Large Language Models Show Amplified Cognitive Biases in Moral Decision-Making},
  author = {Cheung, Vanessa and Maier, Maximilian and Lieder, Falk},
  year = 2025,
  journal = {Proceedings of the National Academy of Sciences},
  volume = {122},
  number = {25},
  eprint = {https://www.pnas.org/doi/pdf/10.1073/pnas.2412015122},
  pages = {e2412015122},
  doi = {10.1073/pnas.2412015122}
}

@inproceedings{daniels-kochExpertiseProblemLearning2022,
  title = {The Expertise Problem: {{Learning}} from Specialized Feedback},
  booktitle = {{{NeurIPS ML}} Safety Workshop},
  author = {{Daniels-Koch}, Oliver and Freedman, Rachel},
  year = 2022
}

@inproceedings{ghosalEffectModelingHuman2022,
  title = {The Effect of Modeling Human Rationality Level on Learning Rewards from Multiple Feedback Types},
  booktitle = {{{AAAI}} Conference on Artificial Intelligence},
  author = {Ghosal, Gaurav R. and Zurek, Matthew and Brown, Daniel S. and Dragan, Anca D.},
  year = 2022
}

@misc{itzhakInstructedBiasInstructiontuned2024,
  title = {Instructed to Bias: {{Instruction-tuned}} Language Models Exhibit Emergent Cognitive Bias},
  author = {Itzhak, Itay and Stanovsky, Gabriel and Rosenfeld, Nir and Belinkov, Yonatan},
  year = 2024,
  eprint = {2308.00225},
  primaryclass = {cs.AI},
  archiveprefix = {arXiv}
}

@article{jeonRewardrationalImplicitChoice2020a,
  title = {Reward-Rational (Implicit) Choice: {{A}} Unifying Formalism for Reward Learning},
  author = {Jeon, Hong Jun and Milli, Smitha and Dragan, Anca D.},
  year = 2020,
  journal = {ArXiv},
  volume = {abs/2002.04833}
}

@incollection{kahnemanThinkingFastSlow2013,
  title = {Thinking, Fast and Slow},
  booktitle = {Thinking, Fast and Slow},
  author = {Kahneman, Daniel},
  year = 2013,
  edition = {First paperback edition.},
  publisher = {{Farrar, Straus and Giroux}},
  address = {New York},
  isbn = {978-0-374-53355-7},
  langid = {english}
}

@misc{knoxModelsHumanPreference2023,
  title = {Models of Human Preference for Learning Reward Functions},
  author = {Knox, W. Bradley and {Hatgis-Kessell}, Stephane and Booth, Serena and Niekum, Scott and Stone, Peter and Allievi, Alessandro},
  year = 2023,
  eprint = {2206.02231},
  primaryclass = {cs.LG},
  archiveprefix = {arXiv}
}

@article{macmillan-scottIrrationalityAIState2023,
  title = {({{Ir}})Rationality in {{AI}}: {{State}} of the Art, Research Challenges and Open Questions},
  author = {{Macmillan-Scott}, Olivia and Musolesi, Mirco},
  year = 2023,
  journal = {Artificial Intelligence Review},
  volume = {58},
  pages = {352}
}

@misc{malbergComprehensiveEvaluationCognitive2024,
  title = {A Comprehensive Evaluation of Cognitive Biases in Llms},
  author = {Malberg, Simon and Poletukhin, Roman and Schuster, Carolin M. and Groh, Georg},
  year = 2024,
  eprint = {2410.15413},
  primaryclass = {cs.CL},
  archiveprefix = {arXiv}
}

@incollection{russellHumanCompatibleAI2019,
  title = {Human Compatible : {{AI}} and the Problem of Control},
  booktitle = {Human Compatible : {{AI}} and the Problem of Control},
  author = {Russell, Stuart J.},
  year = 2019,
  publisher = {Penguin Books},
  address = {London},
  isbn = {978-0-241-33524-6},
  langid = {english}
}

@misc{shahFeasibilityLearningRather2019,
  title = {On the Feasibility of Learning, Rather than Assuming, Human Biases for Reward Inference},
  author = {Shah, Rohin and Gundotra, Noah and Abbeel, Pieter and Dragan, Anca D.},
  year = 2019,
  eprint = {1906.09624},
  primaryclass = {cs.LG},
  archiveprefix = {arXiv}
}

@inproceedings{singhalScalableOversightAccounting,
  title = {Scalable Oversight by Accounting for Unreliable Feedback},
  author = {Singhal, Shivam and Laidlaw, Cassidy and Dragan, Anca},
  year = 2024,
  booktitle = {ICML Workshop on Models of Human Feedback for AI Alignment}
}

@article{boothPerilsTrialanderrorReward2023a,
  title = {The Perils of Trial-and-Error Reward Design: {{Misdesign}} through Overfitting and Invalid Task Specifications},
  author = {Booth, Serena and Knox, W. Bradley and Shah, Julie and Niekum, Scott and Stone, Peter and Allievi, Alessandro},
  year = 2023,
  month = jun,
  journal = {Proceedings of the AAAI Conference on Artificial Intelligence},
  volume = {37},
  number = {5},
  pages = {5920--5929},
  doi = {10.1609/aaai.v37i5.25733}
}

@inproceedings{horterBoothShouldRobotsComply2026,
author = {Horter, Tiffany and Markham, Andrew and Trigoni, Niki and Booth, Serena},
title = {Should Robots Comply with Our Instructions or Intentions?},
year = {2026},
isbn = {9798400721281},
publisher = {Association for Computing Machinery},
address = {New York, NY, USA},
url = {https://doi.org/10.1145/3757279.3785553},
doi = {10.1145/3757279.3785553},
booktitle = {Proceedings of the 21st ACM/IEEE International Conference on Human-Robot Interaction},
pages = {246–254},
numpages = {9},
location = {Edinburgh, Scotland, UK},
series = {HRI '26}
}

@misc{hong2023sensitivityrewardinferencemisspecified,
      title={On the Sensitivity of Reward Inference to Misspecified Human Models}, 
      author={Joey Hong and Kush Bhatia and Anca Dragan},
      year={2023},
      eprint={2212.04717},
      archivePrefix={arXiv},
      primaryClass={cs.LG},
      url={https://arxiv.org/abs/2212.04717}, 
}

@misc{chan2021humanirrationalitybadgood,
      title={Human irrationality: both bad and good for reward inference}, 
      author={Lawrence Chan and Andrew Critch and Anca Dragan},
      year={2021},
      eprint={2111.06956},
      archivePrefix={arXiv},
      primaryClass={cs.LG},
      url={https://arxiv.org/abs/2111.06956}, 
}

@misc{hosking2024humanfeedbackgoldstandard,
      title={Human Feedback is not Gold Standard}, 
      author={Tom Hosking and Phil Blunsom and Max Bartolo},
      year={2024},
      eprint={2309.16349},
      archivePrefix={arXiv},
      primaryClass={cs.CL},
      url={https://arxiv.org/abs/2309.16349}, 
}

@inproceedings{zhong2025balancing,
title     = {Balancing Rigor and Utility: Mitigating Cognitive Biases in Large Language Models for Multiple-Choice Questions},
author    = {Zhong, H. and Wang, L. and Cao, Wenting and Sun, Zeyuan},
booktitle = {Proceedings of the Annual Meeting of the Cognitive Science Society},
volume    = {47},
year      = {2025},
publisher = {Cognitive Science Society},
url       = {https://escholarship.org/uc/item/2vr690cx}
}

@inproceedings{malberg-etal-2025-comprehensive,
    title = "A Comprehensive Evaluation of Cognitive Biases in {LLM}s",
    author = "Malberg, Simon and
    Poletukhin, Roman and
    Schuster, Carolin and
    Groh, Georg Groh",
    editor = {H{\"a}m{\"a}l{\"a}inen, Mika and
    {\"O}hman, Emily and
    Bizzoni, Yuri and
    Miyagawa, So and
    Alnajjar, Khalid},
    booktitle = "Proceedings of the 5th International Conference on Natural Language Processing for Digital Humanities",
    month = may,
    year = "2025",
    address = "Albuquerque, USA",
    publisher = "Association for Computational Linguistics",
    url = "https://aclanthology.org/2025.nlp4dh-1.50/",
    doi = "10.18653/v1/2025.nlp4dh-1.50",
    pages = "578--613",
    ISBN = "979-8-89176-234-3"
}

@incollection{kahneman2013prospect,
  title={Prospect theory: An analysis of decision under risk},
  author={Kahneman, Daniel and Tversky, Amos},
  booktitle={Handbook of the fundamentals of financial decision making: Part I},
  pages={99--127},
  year={2013},
  publisher={World Scientific}
}

@inproceedings{moore2025and,
  title={When and Why Hyperbolic Discounting Matters for Reinforcement Learning Interventions},
  author={Moore, Ian M and Nofshin, Eura and Swaroop, Siddharth and Murphy, Susan and Doshi-Velez, Finale and Pan, Weiwei},
  booktitle={Reinforcement Learning Conference},
  year={2025}
}

@inproceedings{dalonzo2026helpful,
  author = {D'Alonzo, Samantha and Kreuter, Frauke and Booth, Serena},
  title = {Helpful, Harmless, Honest? RLHF as Survey Design and Content Moderation},
  booktitle = {Proceedings of the 2026 ACM Conference on Fairness, Accountability, and Transparency (FAccT '26)},
  year = {2026},
  location = {Montreal, QC, Canada},
  dates = {June 25--28, 2026},
  publisher = {ACM},
  doi = {10.1145/3805689.3806444},
  isbn = {979-8-4007-2596-8/2026/06},
  copyright = {Creative Commons Attribution (CC-BY)}
}

@article{hatgis2025influencing,
  title={Influencing humans to conform to preference models for {RLHF}},
  author={Hatgis-Kessell, Stephane and Knox, W Bradley and Booth, Serena and Niekum, Scott and Stone, Peter},
  journal={arXiv preprint arXiv:2501.06416},
  year={2025}
}

@article{guthrie2007blinking,
  title={Blinking on the bench: How judges decide cases},
  author={Guthrie, Chris and Rachlinski, Jeffrey J and Wistrich, Andrew J},
  journal={Cornell L. Rev.},
  volume={93},
  pages={1},
  year={2007},
  publisher={HeinOnline}
}

@misc{shao2024deepseekmathpushinglimitsmathematical,
      title={DeepSeekMath: Pushing the Limits of Mathematical Reasoning in Open Language Models}, 
      author={Zhihong Shao and Peiyi Wang and Qihao Zhu and Runxin Xu and Junxiao Song and Xiao Bi and Haowei Zhang and Mingchuan Zhang and Y. K. Li and Y. Wu and Daya Guo},
      year={2024},
      eprint={2402.03300},
      archivePrefix={arXiv},
      primaryClass={cs.CL},
      url={https://arxiv.org/abs/2402.03300}, 
}

@article{TverskyAmos1983Evir,
author = {Tversky, Amos and Kahneman, Daniel},
address = {Washington, etc},
title = {Extensional versus intuitive reasoning: The conjunction fallacy in probability judgment},
volume = {90},
year = {1983},
copyright = {1983 American Psychological Association},
issn = {0033-295X},
publisher = {American Psychological Association},
journal = {Psychological review},
language = {eng},
number = {4},
pages = {293-315},
}

@Article{ke2024medicalCogBias,
author="Ke, Yuhe
and Yang, Rui
and Lie, Sui An
and Lim, Taylor Xin Yi
and Ning, Yilin
and Li, Irene
and Abdullah, Hairil Rizal
and Ting, Daniel Shu Wei
and Liu, Nan",
title="Mitigating Cognitive Biases in Clinical Decision-Making Through Multi-Agent Conversations Using Large Language Models: Simulation Study",
journal="J Med Internet Res",
year="2024",
month="Nov",
day="19",
volume="26",
pages="e59439",
issn="1438-8871",
doi="10.2196/59439",
url="https://www.jmir.org/2024/1/e59439",
url="https://doi.org/10.2196/59439"
}

@misc{jiang2023mistral7b,
      title={Mistral 7B}, 
      author={Albert Q. Jiang and Alexandre Sablayrolles and Arthur Mensch and Chris Bamford and Devendra Singh Chaplot and Diego de las Casas and Florian Bressand and Gianna Lengyel and Guillaume Lample and Lucile Saulnier and Lélio Renard Lavaud and Marie-Anne Lachaux and Pierre Stock and Teven Le Scao and Thibaut Lavril and Thomas Wang and Timothée Lacroix and William El Sayed},
      year={2023},
      eprint={2310.06825},
      archivePrefix={arXiv},
      primaryClass={cs.CL},
      url={https://arxiv.org/abs/2310.06825}, 
}

@misc{lin2022truthfulqameasuringmodelsmimic,
      title={TruthfulQA: Measuring How Models Mimic Human Falsehoods}, 
      author={Stephanie Lin and Jacob Hilton and Owain Evans},
      year={2022},
      eprint={2109.07958},
      archivePrefix={arXiv},
      primaryClass={cs.CL},
      url={https://arxiv.org/abs/2109.07958}, 
}

@misc{qwen3technicalreport,
      title={Qwen3 Technical Report}, 
      author={Qwen Team},
      year={2025},
      eprint={2505.09388},
      archivePrefix={arXiv},
      primaryClass={cs.CL},
      url={https://arxiv.org/abs/2505.09388}, 
}

@misc{openai2025gpt52systemcard,
  author       = {{OpenAI}},
  title        = {{Update to GPT-5 System Card: GPT-5.2}},
  year         = {2025},
  howpublished = {\url{https://openai.com/index/gpt-5-system-card-update-gpt-5-2/}},
}

@misc{burns2023weaktostronggeneralizationelicitingstrong,
      title={Weak-to-Strong Generalization: Eliciting Strong Capabilities With Weak Supervision}, 
      author={Collin Burns and Pavel Izmailov and Jan Hendrik Kirchner and Bowen Baker and Leo Gao and Leopold Aschenbrenner and Yining Chen and Adrien Ecoffet and Manas Joglekar and Jan Leike and Ilya Sutskever and Jeff Wu},
      year={2023},
      eprint={2312.09390},
      archivePrefix={arXiv},
      primaryClass={cs.CL},
      url={https://arxiv.org/abs/2312.09390}, 
}

@article{ethayarajh2024kto,
  title={{KTO}: Model alignment as prospect theoretic optimization},
  author={Ethayarajh, Kawin and Xu, Winnie and Muennighoff, Niklas and Jurafsky, Dan and Kiela, Douwe},
  journal={arXiv preprint arXiv:2402.01306},
  year={2024}
}

@inproceedings{knox2024learning,
  title={Learning optimal advantage from preferences and mistaking it for reward},
  author={Knox, W Bradley and Hatgis-Kessell, Stephane and Adalgeirsson, Sigurdur Orn and Booth, Serena and Dragan, Anca and Stone, Peter and Niekum, Scott},
  booktitle={Proceedings of the AAAI Conference on Artificial Intelligence},
  volume={38},
  number={9},
  pages={10066--10073},
  year={2024}
}
\medskip
%%%%%%%%%%%%%%%%%%%%%%%%%%%%%%%%%%%%%%%%%%%%%%%%%%%%%%%%%%%%

\appendix
\newpage 
\section{$\beta_{new}$ formula ablation}
\label{app:beta-formula-ablation-scenario}
As an alternative to equation \ref{eqn:b_new_main}, we instead  considered the following formula for setting $\beta_{new}$:
\[\beta_{new} = \texttt{logistic}(k * D_s * (c - D_f)) \]
where $D_s$ is the likelihood from the bias detector that the situation is one in which bias may be present, and $D_f$ is the likelihood that the given feedback is biased. This uses a logit transform to make it smooth around the ends. However, we determined using an ablation study on the BRU dataset (shown in Table \ref{tab:mistake_fn_ablation_d1}) that the inclusion of the $D_s$ factor in fact degraded performance. This is likely because the scenario is essentially repeated as context for the bias detector in the $D_f$ factor but lacks the information about whether a biased choice was actually made or not; as discussed throughout the text, if a preference reflects a rational choice in the face of bias, that information is highly useful for learning and should not be discarded through a low $\beta$. Using the $D_s$ factor only while holding the $D_f$ factor constant at the pooled mean of $D_f$ made it collapse to near the baseline, indicating the potential weakness of a $\beta$ that relies only on the scenario. 

\begin{table}[h]                                                                                              
  \centering          
  \caption{Mistake-function ablation on BRU dataset. ``Full'' uses both \(s\) (scenario) and \(r\) (response) bias scores; ``response only'' holds \(s{=}1\); ``scenario only'' pools gt/cb response scores to their mean. Diff and \(p\) are vs.\ the GRPO baseline (\(\beta{=}1.0\), GT~\(=9.9\%\)); Cohen's \(d\) is paired across test prompts; \(p\) is from a \(10\,000\)-sample paired bootstrap.} 
  \label{tab:mistake_fn_ablation_d1}    
  \begin{tabular}{lrrrrrr}              
  \toprule                                        
  Variant & GT\,\% & CB\,\% & NA\,\% & $\Delta$\,GT & Cohen $d$ & $p$ \\    
  \midrule                                                                                                     
  Both                      & 50.2 & 47.1 & 2.7 & $+40.3$ & 0.99 & $<\!0.001$ \\  
  Response only ($s{=}1$)   & 54.6 & 43.4 & 2.0 & $+44.7$ & 1.30 & $<\!0.001$ \\                                
  Scenario only (pooled $r$) &  9.4 & 89.4 & 1.1 & $-0.5$  & $-0.11$ & $0.716$ \\                               
  \bottomrule   
  \end{tabular}    
  \end{table}

\section{Impact of Debiasing by Bias Type}
Here we present the results of the debiased model and consider the effect on each bias type. 95$\%$ CI width is reported. 

\begin{figure}[h]
    \centering
    \includegraphics[width=0.5\linewidth]{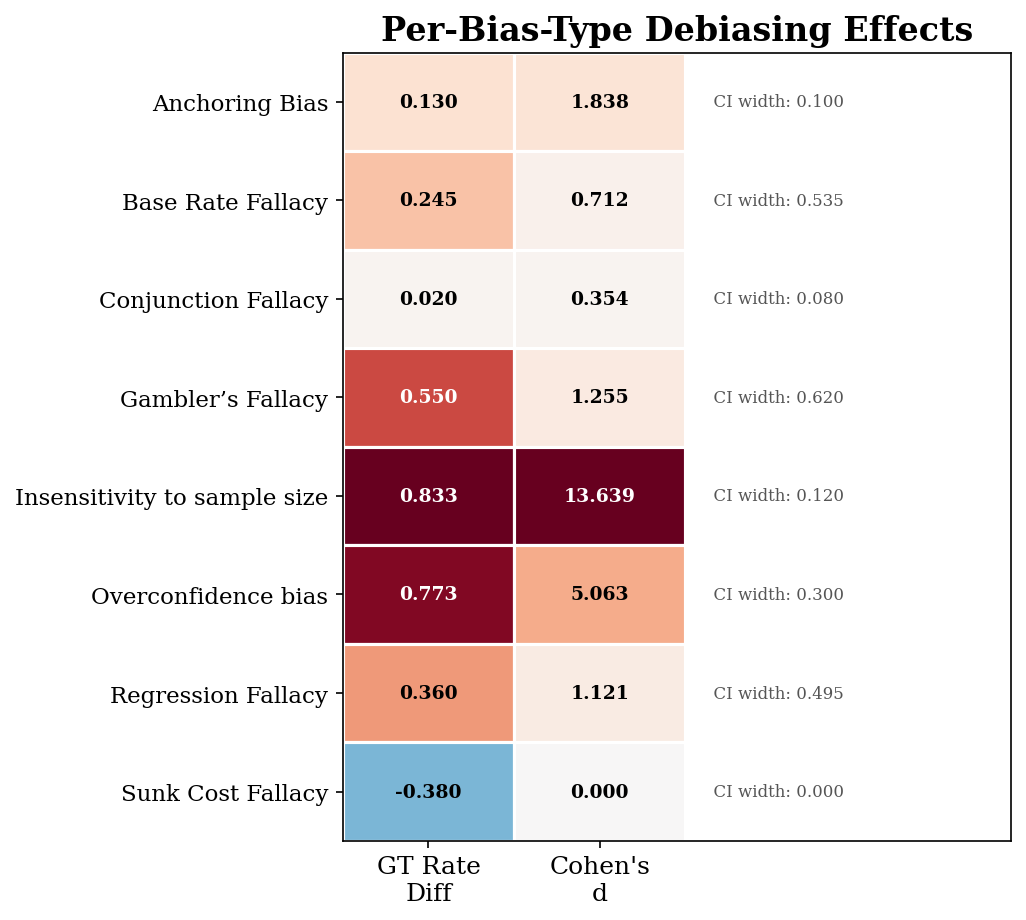}
    \caption{Heat map of debiasing effects by bias type for BRU dataset.}
    \label{fig:bias-type-table-d1}
\end{figure}

\begin{figure}[h]
    \centering
    \includegraphics[width=1.0\linewidth]{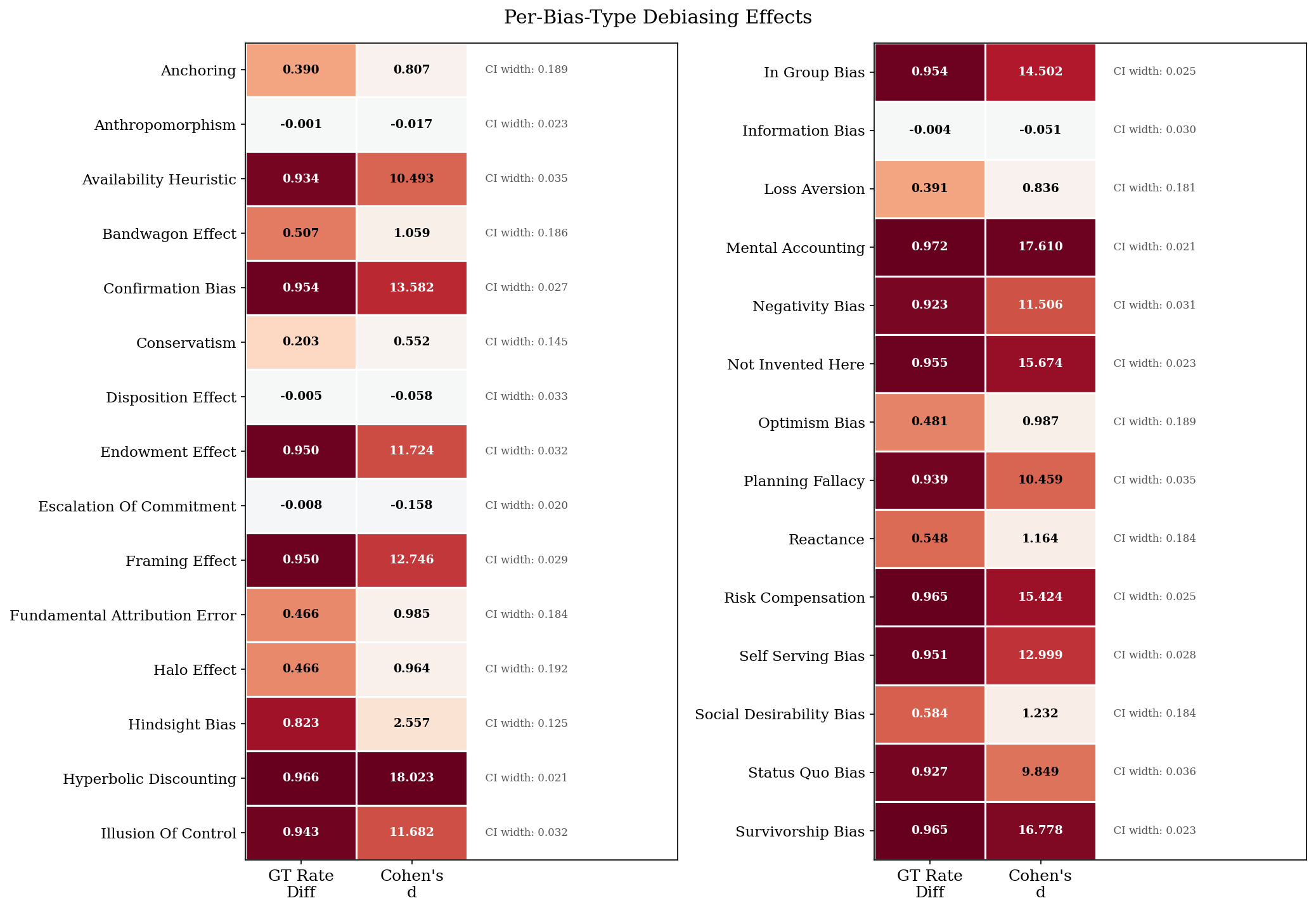}
    \caption{Heat map of debiasing effects by bias type for CogBias dataset.}
    \label{fig:bias-type-table-d2}
\end{figure}

\newpage
\section{Training hyperparameters for CogBias Dataset}
On publication, we will release the code we used to train the models. 
\label{app:cogbias-training-hyperparams}

\providecommand{\code}[1]{\texttt{\small #1}}
\providecommand{\file}[1]{\texttt{\small #1}}

All values reflect \file{config/judge.yaml} and the source files under \file{src/}. 

% ---------------------------------------------------------------------------
\begin{table}[h]
\centering
\resizebox{\linewidth}{!}{%
\begin{tabular}{p{5cm} p{9cm}}
\toprule
\textbf{Setting} & \textbf{Value} \\
\midrule
Base model & \code{mistralai/Mistral-7B-v0.1} \\
\code{torch\_dtype} & \code{bfloat16} \\
Attention impl. & \code{flash\_attention\_2} (CUDA); SDPA fallback (MPS/CPU) \\
RM quant.\ during GRPO & 4-bit (\code{bnb\_4bit\_compute\_dtype=bf16}) \\
Policy quant.\ during GRPO & full bf16 (4-bit only if \code{model.use\_4bit=true}) \\
\bottomrule
\end{tabular}}
\caption{CogBias: Base-model loading and numeric precision.}
\end{table}

% ---------------------------------------------------------------------------
\begin{table}[h]
\centering
\resizebox{\linewidth}{!}{%
\begin{tabular}{p{5cm} p{9cm}}
\toprule
\textbf{Setting} & \textbf{Value} \\
\midrule
\code{r} & 16 \\
\code{lora\_alpha} & 32 \\
\code{lora\_dropout} & 0.05 \\
\code{target\_modules} & \code{q\_proj, k\_proj, v\_proj, o\_proj} \\
\code{bias} & \code{none} \\
\code{task\_type} & \code{SEQ\_CLS} (RM) / \code{CAUSAL\_LM} (policy) \\
\code{modules\_to\_save} & none \\
\bottomrule
\end{tabular}}
\caption{CogBias: LoRA adapter configuration. Identical for the reward model and the policy except for \code{task\_type}.}
\end{table}

% ---------------------------------------------------------------------------
\begin{table}[h]
\centering
\resizebox{\linewidth}{!}{%
\begin{tabular}{p{5cm} p{9cm}}
\toprule
\textbf{Setting} & \textbf{Value} \\
\midrule
Format & \code{dataset2} (control/treatment, 29 bias types) \\
Splits & \code{data/dataset2\_\{train,val,test\}.csv} \\
\code{option\_mode} & \code{ab} (binary A/B reduction) \\
\code{stratify\_by} & \code{bias\_type} \\
\code{train\_sample\_frac} & 1.0 (no downsampling) \\
\code{reduce\_test} & \code{true} \\
Shuffle / A--B alternation seed & 42 \\
Stratified-downsample seed & 42 (used only when \code{train\_sample\_frac<1.0}) \\
Score calibration & applied at load time from \code{outputs/judge/score\_calibration.json} if present \\
\bottomrule
\end{tabular}}
\caption{CogBias: Data loading.}
\end{table}

% ---------------------------------------------------------------------------
\begin{table}[p]
\centering
\resizebox{\linewidth}{!}{%
\begin{tabular}{p{5cm} p{9cm}}
\toprule
\textbf{Setting} & \textbf{Value} \\
\midrule
Formula & $\beta = \sigma\!\bigl(\mathrm{rc}\,\cdot\,(\tau - s)\bigr)$, $s$ = per-response bias score \\
\code{k} (rc in code) & 30.0 \\
\code{threshold} ($\tau$) & 0.40 (\code{auto} falls back to pooled-median) \\
Range & $\beta \in [0,1]$, no clamping \\
Baseline-branch override & $\beta = 1.0$ for every pair \\
Bias-score source & per-example LLM-judge scores in CSV \\
\bottomrule
\end{tabular}}
\caption{CogBias - Mistake function (dynamic $\beta$, debiased branch only).}
\end{table}

% ---------------------------------------------------------------------------
\begin{table}[p]
\centering
\resizebox{\linewidth}{!}{%
\begin{tabular}{p{5cm} p{9cm}}
\toprule
\textbf{Setting} & \textbf{Value} \\
\midrule
\code{pair\_sampling} & \code{fixed} (default) \\
\code{gt\_pairs} (correct) & 1 per question \\
\code{cb\_pairs} (biased) & 3 per question \\
Effective ratio (correct:biased) & 1:3 \\
Random-mode seed (if enabled) & 42 (validation: 43) \\
Suffix appended to prompt & \code{"$\backslash$nChoose A or B.$\backslash$nAnswer: "} \\
\bottomrule
\end{tabular}}
\caption{CogBias - Pair generation per training row. }
\end{table}

% ---------------------------------------------------------------------------
\begin{table}[p]
\centering
\resizebox{\linewidth}{!}{%
\begin{tabular}{p{5cm} p{9cm}}
\toprule
\textbf{Setting} & \textbf{Value} \\
\midrule
Loss & $-\log \sigma\bigl(\beta\cdot(r_{\text{chosen}}-r_{\text{rejected}})\bigr)$ \\
Epochs & 3 \\
Per-device batch size & 8 \\
\code{gradient\_accumulation\_steps} & 4 (inherits \code{training.*}) \\
Effective batch (single-GPU) & 32 \\
Learning rate & $1\times 10^{-5}$ \\
\code{max\_length} & 512 \\
\code{gradient\_checkpointing} & \code{true} \\
\code{bf16} / \code{fp16} & auto from \code{torch\_dtype} (bf16=true) \\
\code{save\_strategy}/\code{\_steps}/\code{\_total\_limit} & \code{steps} / 50 / 2 \\
\code{remove\_unused\_columns} & \code{true} \\
\code{center\_rewards\_coefficient} & \code{None} (term disabled) \\
Eval-set cap & first 200 validation samples \\
\code{report\_to} & \code{tensorboard} \\
\bottomrule
\end{tabular}}
\caption{CogBias: Reward-model training arguments. Both baseline ($\beta=1$) and debiased branches share these; only the per-row $\beta$ column differs.}
\end{table}

% ---------------------------------------------------------------------------
\begin{table}[p]
\centering
\resizebox{\linewidth}{!}{%
\begin{tabular}{p{5cm} p{9cm}}
\toprule
\textbf{Setting} & \textbf{Value} \\
\midrule
Procedure & mean RM logit over \code{prompt+suffix+gt} and \code{prompt+suffix+cb} for every train row \\
Batch size & 8 \\
Tokenization \code{max\_length} & 512 \\
Cache & \code{<rm\_path>/calibration.json} (skipped if present) \\
Used as & \code{reward = (rm\_score $-$ offset) + format\_bonus} \\
\bottomrule
\end{tabular}}
\caption{CogBias - Reward-model offset calibration (pre-GRPO). Zero-centering offset computed once per RM and cached; ensures the GRPO advantage term is not biased by the RM's mean logit.}
\end{table}

% ---------------------------------------------------------------------------
\begin{table}[p]
\centering
\resizebox{\linewidth}{!}{%
\begin{tabular}{p{5cm} p{9cm}}
\toprule
\textbf{Setting} & \textbf{Value} \\
\midrule
\code{num\_generations} per prompt & 4 \\
\code{max\_completion\_length} & 15 tokens \\
\code{temperature} & 0.7 \\
\code{beta} (KL coef.\ vs ref.\ policy) & 0.04 \\
\code{num\_iterations} & 1 \\
\code{loss\_type} & \code{grpo} \\
Learning rate & $1\times 10^{-5}$ \\
Per-device batch size & 4 \\
\code{gradient\_accumulation\_steps} & 4 \\
Effective batch (single-GPU) & 16 \\
Epochs & 2 \\
\code{gradient\_checkpointing} & \code{true} \\
\code{train\_order} & \code{debiased\_first} \\
Tokenizer \code{padding\_side} & \code{left} (forced for generation) \\
Seed & 42 (default arg of \code{\_grpo\_finetune}) \\
\code{save\_strategy}/\code{\_steps}/\code{\_total\_limit} & \code{steps} / 50 / 2 \\
\code{report\_to} & \code{tensorboard} \\
\bottomrule
\end{tabular}}
\caption{CogBias - GRPO trainer configuration.}
\end{table}

% ---------------------------------------------------------------------------
\begin{table}[p]
\centering
\resizebox{\linewidth}{!}{%
\begin{tabular}{p{5cm} p{9cm}}
\toprule
\textbf{Setting} & \textbf{Value} \\
\midrule
Answer regex (AB mode) & \code{[ABab]} \\
Reward (parseable answer) & \code{(rm\_logit $-$ offset) + format\_reward\_bonus} \\
Reward (unparseable) & \code{$-$format\_reward\_bonus} \\
\code{format\_reward\_bonus} & 2.0 \\
RM scoring \code{max\_length} & 512 (truncation, dynamic padding) \\
Clipping / normalization & none \\
\bottomrule
\end{tabular}}
\caption{CogBias - GRPO Reward function}
\end{table}

% ---------------------------------------------------------------------------
\begin{table}[p]
\centering
\resizebox{\linewidth}{!}{%
\begin{tabular}{p{5cm} p{9cm}}
\toprule
\textbf{Setting} & \textbf{Value (HF default; not overridden)} \\
\midrule
Optimizer & \code{adamw\_torch} \\
\code{lr\_scheduler\_type} & \code{linear} \\
\code{warmup\_ratio} / \code{warmup\_steps} & 0 / 0 \\
\code{weight\_decay} & 0.0 \\
\code{max\_grad\_norm} & 1.0 \\
AdamW $(\beta_1, \beta_2)$ / $\epsilon$ & $(0.9, 0.999)$ / $1\times 10^{-8}$ \\
\bottomrule
\end{tabular}}
\caption{CogBias - Trainer defaults inherited from \texttt{transformers.TrainingArguments}; not set in \file{config/judge.yaml}.}
\end{table}

% ---------------------------------------------------------------------------
\subsection{Distributed-training scaling}
When the pipeline is launched under accelerate / torch DDP with
\code{WORLD\_SIZE>1} (e.g.\ \file{scripts/run\_grpo\_parallel.py}),
\code{maybe\_scale\_batch\_for\_ddp} (\file{main.py:42--55}) preserves the
single-GPU effective batch size by:
\begin{enumerate}
  \item dividing \code{gradient\_accumulation\_steps} by \code{WORLD\_SIZE}
        when cleanly divisible (memory-neutral); otherwise
  \item dividing per-device \code{batch\_size} by \code{WORLD\_SIZE}.
\end{enumerate}
The rule is applied to both the \code{training.*} (GRPO) and
\code{reward\_model.*} sections so that ablations launched on different machines remain comparable.

% ---------------------------------------------------------------------------
\subsection{Evaluation}
\begin{table}[p]
\centering
\resizebox{\linewidth}{!}{%
\begin{tabular}{p{5cm} p{9cm}}
\toprule
\textbf{Setting} & \textbf{Value} \\
\midrule
\code{logging.log\_every\_n\_steps} & 10 \\
\code{logging.tensorboard} & \code{true} \\
\code{evaluation.n\_samples} & 10 generations / prompt \\
\code{evaluation.max\_new\_tokens} & 15 \\
\bottomrule
\end{tabular}}
\caption{CogBias - Logging and post-training evaluation}
\end{table}

%%%%%%%%%%%%%%%%%%%%%%%%%%%%%%%%%%%%%%%%%%%%%%%%%%%%%%%%%%%%
\newpage
\section{Training hyperparameters for BRU Dataset}
\label{app:bru-training-hyperparameters}
BRU dataset is run via \file{config/default.yaml}:

All settings not listed below match the CogBias dataset tables.

% ---------------------------------------------------------------------------
\begin{table}[h]
\centering
\resizebox{\linewidth}{!}{%
\begin{tabular}{p{5cm} p{9cm}}
\toprule
\textbf{Setting} & \textbf{Value} \\
\midrule
Source & single CSV \code{data/BRUdataset3.csv} \\
Split policy & stratified \code{train\_test\_split} (seed 42) on \code{bias\_type} \\
\code{train} / \code{val} / \code{test} ratio & 0.8 / 0.1 / 0.1 \\
Score calibration applied & no -- \code{\_apply\_score\_calibration} is dataset~2 only \\
Judge held-out exclusion & not applied - no calibration\\
\bottomrule
\end{tabular}}
\caption{BRU dataset - Data}
\end{table}

% ---------------------------------------------------------------------------
\begin{table}[h]
\centering
\resizebox{\linewidth}{!}{%
\begin{tabular}{p{5cm} p{9cm}}
\toprule
\textbf{Setting} & \textbf{Value} \\
\midrule
\code{rationality\_constant} (rc) & 100.0 \\
\code{threshold} ($\tau$) & \code{auto} -- median of all training response bias scores (label-blind) \\
\bottomrule
\end{tabular}}
\caption{BRU dataset - Mistake function}
\end{table}

% ---------------------------------------------------------------------------
\begin{table}[h]
\centering
\resizebox{\linewidth}{!}{%
\begin{tabular}{p{5cm} p{9cm}}
\toprule
\textbf{Setting} & \textbf{Value} \\
\midrule
Epochs & 4 \\
Per-device batch size & 4 \\
\code{gradient\_accumulation\_steps} & 1 \\
Effective batch (single-GPU) & 4 \\
\bottomrule
\end{tabular}}
\caption{BRU dataset - Reward-model training.}
\end{table}

% ---------------------------------------------------------------------------
\begin{table}[h]
\centering
\resizebox{\linewidth}{!}{%
\begin{tabular}{p{5cm} p{9cm}}
\toprule
\textbf{Setting} & \textbf{Value} \\
\midrule
Per-device batch size & 16 \\
\code{gradient\_accumulation\_steps} & 1 \\
Effective batch (single-GPU) & 16 \\
\code{num\_generations} per prompt & 8 \\
\code{evaluation.n\_samples} & 50 generations / prompt \\
\bottomrule
\end{tabular}}
\caption{BRU dataset - GRPO finetuneing and evaluation}
\end{table}

%%%%%%%%%%%%%%%%%%%%%%%%%%%%%%%%%%%%%%%%%%%%%%%%%%%%%%%%%%%%%%%%
\newpage
\section{Robustness hyperparameters}
\label{app:robustness-hyperparam}
For Qwen3-8B-Base run: 
\begin{verbatim}
    "model": {"use_4bit": True},
    "training": {"batch_size": 1, "gradient_accumulation_steps": 16},
    "grpo": {"num_generations": 8, "temperature": 1.0},
\end{verbatim}

\section{Compute Resources}
All experiments were able to be run on a maximum of 4 A10s. To reproduce an experiment where a model passes through the reward model training and then GRPO from end to end to be finetuned would take approximately 36 hours on 2 A10s. Multiply this number by the number of variants present in the experiment for an estimate of wall-clock time per experiment. 
\newpage

\end{document}